\documentclass{article}
\usepackage{graphicx} 
\usepackage{amsfonts}
\usepackage{amsmath}
\usepackage{subcaption}
\usepackage{hyperref}
\usepackage{authblk}

\title{Stochastic Optimal Control via Measure Relaxations}
\author{Etienne Buehrle}
\author{Christoph Stiller}
\affil[]{Karlsruhe Institute of Technology\\\texttt{\{etienne.buehrle, stiller\}@kit.edu}}
\date{}

\begin{document}

\maketitle

\begin{abstract}
The optimal control problem of stochastic systems is commonly solved via robust or scenario-based optimization methods, which are both challenging to scale to long optimization horizons. We cast the optimal control problem of a stochastic system as a convex optimization problem over occupation measures, extending \cite{lasserre2008nonlinear}. We demonstrate our method on a set of synthetic and real-world scenarios, learning cost functions from data via Christoffel polynomials. The code for our experiments is available at \url{https://github.com/ebuehrle/dpoc}.
\end{abstract}

\section{Introduction}
The optimal control problem of stochastic systems is commonly solved via robust \cite{bansal2017hamilton, wetzlinger2023fully} or scenario-based \cite{de2005stochastic, shen2021incremental, micheli2022scenario} optimization methods, which are both challenging to scale to long optimization horizons due to their open-loop nature.

Dynamic programming formulations \cite{bertsekas2012dynamic}, while applicable to stochastic systems, typically involve nonconvex optimization problems and do not support specifying the terminal distribution. 

Polynomial optimization has been proposed for deterministic nonlinear \cite{lasserre2008nonlinear} and hybrid systems \cite{marcucci2024graphs}. We extend the method to stochastic systems using a weak formulation of the Fokker-Planck equation. As a cost function, we propose to use the Christoffel polynomial, which can be estimated from data.

\section{Background}
We recall weak optimal control and its connection with optimal transport for a system with dynamics $dx = f(x,u)\,dt$.

\subsection{Optimal Transport}
The task of optimally transporting a measure $\rho_0$ to a measure $\rho_T$ can be formulated as a finite-horizon optimal control problem $\int_0^T\int_{\mathbb{R}^d}\rho(t,x)c(t,x)dxdt$ subject to a continuity equation $\partial_t\rho + \nabla\cdot(f\rho)=0$ and boundary constraints $\rho(0,\cdot)=\rho_0$, $\rho(T,\cdot)=\rho_T$ \cite{benamou2000computational}.

\subsection{Weak Optimal Control}
Weak optimal control defines a measure supported on a time interval $\int_{[0,T]\times{\mathbb{R}^d}} c\,d\rho$, subject to a transport equation $\mathcal{L}^*\rho=\rho_0 - \rho_T$ with $\mathcal{L}^*\rho = -\partial_t\rho - \nabla\cdot(f\rho)$, yielding a linear program in the space of measures that can be canonically transformed into a semidefinite program via the Moment-SOS hierarchy, which can be shown generically to have finite convergence to the global optimum with increasing relaxation degree \cite{lasserre2009moments, weisser2019polynomial}.

\subsection{Dual Polynomial Program}
The dual is a polynomial program \begin{align}
    \max \quad &\langle V, \rho_0 \rangle - \langle V, \rho_T \rangle \\
    \text{s.t.} \quad &\mathcal{L}V \geq -c
\end{align} with $\mathcal{L}V = \partial_tV + \nabla V^\top f$, which can be canonically transformed into a semidefinite program via the Moment-SOS hierarchy \cite{blekherman2012semidefinite, legat2017sos}. The dual variable $V$ can be interpreted as a suboptimal value function \cite{lasserre2009moments}.

\section{Method}
We optimize a polynomial performance criterion subject to a weak formulation of the Fokker-Planck equation for stochastic dynamics $dx = f(x,u)\,dt + \sigma\,dB$.

\subsection{Weak Stochastic Control}
For stochastic dynamics, the state occupation measure evolves according to a Fokker-Planck equation with a drift and a diffusion term $\mathcal{L}^*\rho = -\partial_t\rho - \nabla\cdot(f\rho) + \frac{1}{2}\operatorname{Tr}(\sigma\sigma^\top \nabla^2\rho)$ where $ \nabla^2\rho = (\partial^2\rho/\partial_i\partial_j)_{ij}$ denotes the Hessian \cite{pavliotis2014stochastic}.

\subsection{Dual Polynomial Program}
The Laplacian being self-adjoint, the corresponding dual program is \begin{align}
    \max \quad &\langle V, \rho_0 \rangle - \langle V, \rho_T \rangle \\
    \text{s.t.} \quad &\mathcal{L}V \geq -c
\end{align} where $\mathcal{L}V = \partial_tV + \nabla V^\top f + \frac{1}{2}\operatorname{Tr}(\sigma\sigma^\top \nabla^2 V)$.

\section{Results}
We demonstrate our method on a synthetic constrained stochastic optimal control problem and an optimal control scenario with a polynomial cost function\footnote{Code for our experiments is available at \url{https://github.com/ebuehrle/dpoc}.}.

\subsection{Constrained Stochastic Control}
\begin{figure}
    \centering
    \begin{subfigure}{0.45\linewidth}
        \centering
        \includegraphics[width=\linewidth]{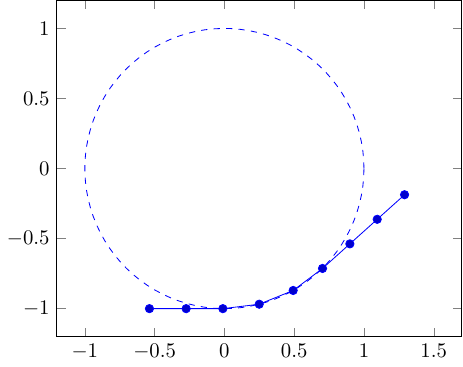}
        \caption{$\frac{1}{2}\sigma\sigma^\top=0$}
        \label{fig:diffusion-means-deterministic}
    \end{subfigure}
    \hfill
    \begin{subfigure}{0.45\linewidth}
        \centering
        \includegraphics[width=\linewidth]{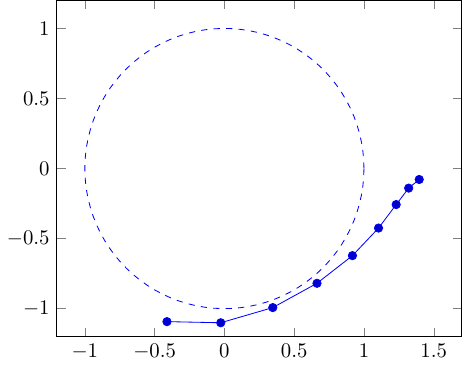}
        \caption{$\frac{1}{2}\sigma\sigma^\top=0.3I$}
        \label{fig:diffusion-means-stochastic}
    \end{subfigure}
    \caption{Stochastic control around an obstacle in the deterministic case (a) and noisy case (b). In the noisy case, the closed-loop nature of the optimization trades off mean and variance.}
    \label{fig:diffusion-means}
\end{figure}

We consider stochastic control with state constraints (Figure~\ref{fig:diffusion-means}), comparing the deterministic (noise-free) and stochastic (noisy) case. In the deterministic case (Figure~\ref{fig:diffusion-means-deterministic}), the expected performance is optimized, resulting in vanishing constraint slackness. In the stochastic case (Figure~\ref{fig:diffusion-means-stochastic}), the closed-loop nature of the optimization trades off mean and variance.

\begin{figure}
    \centering
    \begin{subfigure}{0.45\linewidth}
        \centering
        \includegraphics[width=\linewidth]{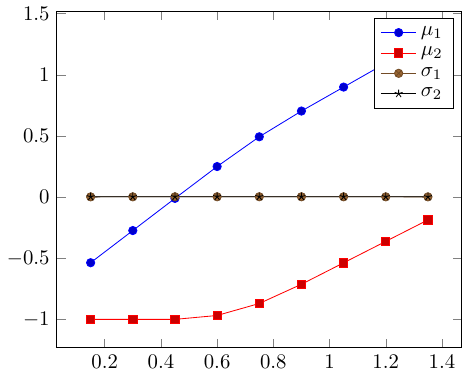}
        \caption{$\frac{1}{2}\sigma\sigma^\top=0$}
        \label{fig:diffusion-deterministic}
    \end{subfigure}
    \hfill
    \begin{subfigure}{0.45\linewidth}
        \centering
        \includegraphics[width=\linewidth]{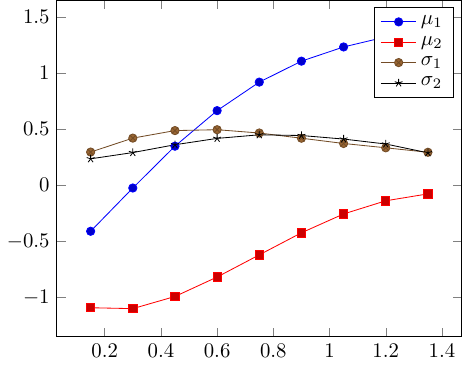}
        \caption{$\frac{1}{2}\sigma\sigma^\top=0.3I$}
        \label{fig:diffusion-stochastic}
    \end{subfigure}
    \caption{Statistics of the occupation measures in Figure \ref{fig:diffusion-means} in the deterministic case (a) and noisy case (b). In the stochastic case, the variance is reduced in radial direction.}
    \label{fig:diffusion}
\end{figure}

The optimization over state-action occupation measures considers the possibility of feedback, avoiding overly conservative assumptions on the system behavior. We plot first and second order statistics of the optimized measures in Figure~\ref{fig:diffusion}. In the deterministic case, the measure is singular with vanishing variance (Figure~\ref{fig:diffusion-deterministic}). In the stochastic case, the control effort is traded off with the variance (risk). In particular, the feedback is used to reduce the variance in the radial direction, leading to fluctuating variance profiles over time (Figure~\ref{fig:diffusion-stochastic}).

\subsection{Polynomial Stochastic Control}
\begin{figure}
    \centering
    \begin{subfigure}{0.32\linewidth}
        \centering
        \includegraphics[width=\linewidth]{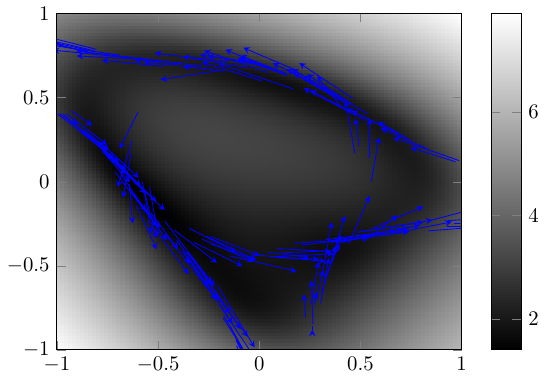}
        \caption{Christoffel function}
        \label{fig:roundabout-christoffel}
    \end{subfigure}
    \hfill
    \begin{subfigure}{0.32\linewidth}
        \centering
        \includegraphics[width=\linewidth]{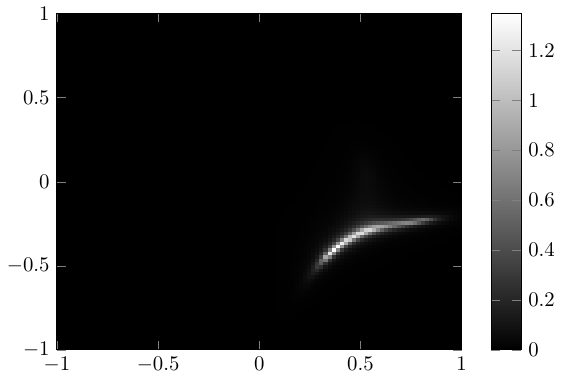}
        \caption{$\frac{1}{2}\sigma\sigma^\top=0$}
        \label{fig:roundabout-deterministic}
    \end{subfigure}
    \hfill
    \begin{subfigure}{0.32\linewidth}
        \centering
        \includegraphics[width=\linewidth]{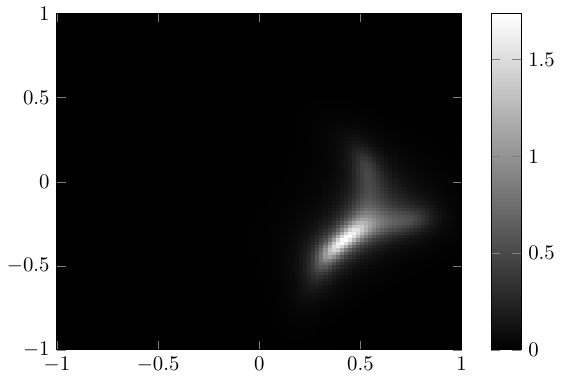}
        \caption{$\frac{1}{2}\sigma\sigma^\top=0.03I$}
        \label{fig:roundabout-stochastic}
    \end{subfigure}
    \caption{Stochastic control with polynomial costs (a). In the deterministic case (b), the measure concentrates on the most likely mode. In the noisy case (c), the measure covers less likely modes.}
    \label{fig:roundabout}
\end{figure}

We consider stochastic control with polynomial costs (Figure~\ref{fig:roundabout}). As a cost function, we propose to use the Christoffel polynomial, which is defined as $\Lambda(x) = \phi(x)^\top M^{-1}\phi(x)$ for a basis\footnote{We use the monomial basis for our experiments.} $\phi$, where $M=\frac{1}{N}\sum_{i=1}^N \phi(x_i)\phi(x_i)^\top$ is an empirical correlation matrix of samples $x_1,\dots,x_N$ \cite{lasserre2022christoffel, henrion2023moments}. An example cost function for a subset of samples from the INTERACTION Dataset of urban driving scenarios \cite{interactiondataset} is shown in Figure~\ref{fig:roundabout-christoffel}.

We consider double integrator (point mass) dynamics with two position and two velocity states, fixing the initial measure at $\delta_{(0.3,-0.5,0.0,0.0)}$. In the deterministic case, the occupation measure follows the empirical flow of the system, concentrating on the predominant mode (Figure~\ref{fig:roundabout-deterministic}). In the stochastic case, the measure exhibits higher entropy, covering less likely modes (Figure~\ref{fig:roundabout-stochastic}).

\subsection{Stochastic Optimal Transport}
\begin{figure}
    \centering
    \includegraphics[width=0.45\linewidth]{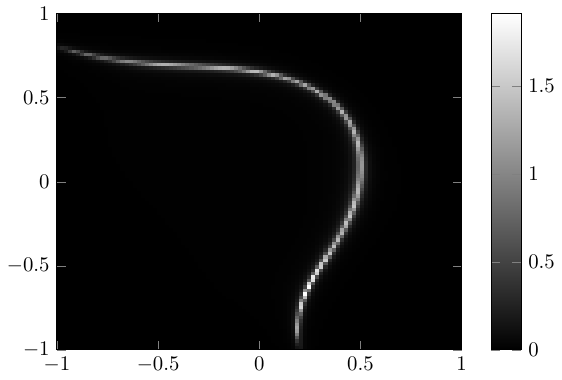}
    \hfill
    \includegraphics[width=0.45\linewidth]{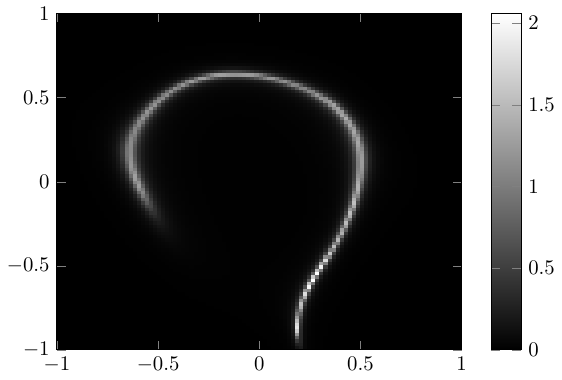}
    \caption{Stochastic optimal transport for two different terminal measures using the cost function in Figure \ref{fig:roundabout-christoffel}. The support of the optimized measure is a subset of the support of the empirical measure, with counter-clockwise flow.}
    \label{fig:roundabout-fixed}
\end{figure}

Finally, we consider the stochastic optimal transport problem with fixed boundary measures and polynomial costs, using the same cost function as in Figure~\ref{fig:roundabout-christoffel}. Figure~\ref{fig:roundabout-fixed} shows results for two different terminal measures. In both cases, the occupation measure follows the empirical flow of the system, which is contained in the cost function.

\section{Conclusion}
We extend the weak optimal transport framework to stochastic systems, finding that the closed-loop nature scales well to long optimization horizons while covering settings including polynomial cost functions and variable horizons.

\subsection{Limitations}
The approach inherits the limitations of semidefinite programming, which is among the most challenging classes of convex optimization problems. Thus, our method is currently not real-time capable. This can be alleviated by methods considering problem structure \cite{coey2022performance, Garstka_2021}, symmetries \cite{gatermann2004symmetry, augier2023symmetry}, or sparsity \cite{magron2023sparse}. We leave these avenues to future work.

\bibliographystyle{plain}
\bibliography{main}

\end{document}